\documentclass[twoside,11pt]{article}

\usepackage[abbrvbib,preprint]{jmlr2e}
\usepackage{multirow}
\usepackage{booktabs}
\usepackage{pifont}
\usepackage{threeparttable}
\usepackage{tcolorbox}
\usepackage{lastpage}
\usepackage{listings}
\usepackage{subfigure}

\definecolor{dkred}{rgb}{0.5,0,0}
\definecolor{dkgreen}{rgb}{0,0.6,0}
\definecolor{gray}{rgb}{0.5,0.5,0.5}
\definecolor{mauve}{rgb}{0.58,0,0.82}

\ShortHeadings{Imbens: Ensemble Class-imbalanced Learning in Python}{Liu et al.}

\firstpageno{1}

\begin{document}

% \title{IMBENS: A Python Toolbox for Ensemble Learning from Class-imbalanced Data}
\title{IMBENS: Ensemble Class-imbalanced Learning in Python}

% \author{\name Zhining Liu \email liu326@illinois.edu\AND
%     \addr University of Illinois at Urbana-Champaign\thanks{Work initialized while at Jilin University.}\\
%     \addr Urbana, IL 61801, USA\\
% }
\author{\name Zhining Liu\textsuperscript{1}\thanks{Work initialized while at Jilin University.} \email liu326@illinois.edu\\
  \name Jian Kang\textsuperscript{1} \email jiank2@illinois.edu\\
  \name Hanghang Tong\textsuperscript{1} \email htong@illinois.edu\\
  % \name Zhepei Wei\textsuperscript{2} \email weizp19@mails.jlu.edu.cn\\
  % \name Erxin Yu\textsuperscript{2} \email yuex19@mails.jlu.edu.cn\\
  % \name Qiang Huang\textsuperscript{2} \email huangqiang18@mails.jlu.edu.cn\\
  % \name Kai Guo\textsuperscript{2} \email guokai20@mail.jlu.edu.cn\\
  % \name Hangting Ye\textsuperscript{2} \email yeht2118@mails.jlu.edu.cn\\
  \name Yi Chang\textsuperscript{2} \email yichang@jlu.edu.cn\\
  \addr \textsuperscript{1} University of Illinois at Urbana-Champaign, Urbana, IL 61801, USA\\
  \addr \textsuperscript{2} Jilin University, Changchun, Jilin 130012, China\\
}
% \editor{Kevin Murphy and Bernhard Sch{\"o}lkopf}
\maketitle

\begin{abstract}%   <- trailing '%' for backward compatibility of .sty file
  % {\tt imbalanced-ensemble}, abbreviated as {\tt imbens}, is an open-source Python toolbox  for quick implementing and deploying ensemble learning algorithms on class-imbalanced data.
  {\tt imbalanced-ensemble}, abbreviated as {\tt imbens}, is an open-source Python toolbox for leveraging the power of ensemble learning to address the class imbalance problem.
  It provides standard implementations of popular ensemble imbalanced learning (EIL) methods with extended features and utility functions.
  These ensemble methods include resampling-based, e.g., under/over-sampling, and reweighting-based, e.g., cost-sensitive learning.
  % Beyond the implementation, we also extend conventional binary EIL algorithms with new functionalities like customizable resampling scheduler, thereby enabling them to handle more complex tasks.
  % Beyond the implementation, we empower EIL algorithms with new functionalities like customizable resampling scheduler and verbose logging, thus enabling them to handle more complex tasks.
  Beyond the implementation, we empower EIL algorithms with new functionalities like customizable resampling scheduler and verbose logging, thus enabling more flexible training and evaluating strategies.
  The package was developed under a simple, well-documented API design that follows {\tt scikit-learn} for increased ease of use.
    {\tt imbens} is released under the MIT open-source license and can be installed from
  Python Package Index (PyPI) or \href{https://github.com/ZhiningLiu1998/imbalanced-ensemble}{\tt https://github.com/ZhiningLiu1998/imbalanced-ensemble}.
  % {\tt imbens} is released under the MIT open-source license and can be installed from 
  % Python Package Index (PyPI).
  % Source code, binaries, detailed documentation, and usage examples are available at 
  % \href{https://github.com/ZhiningLiu1998/imbalanced-ensemble}{\tt https://github.com/ZhiningLiu1998/imbalanced-ensemble}.
\end{abstract}


\section{Introduction}

Class imbalance, also known as the long-tail problem, is the fact that the classes are not represented equally in a classification problem.
Such issue widely exists in many real-world applications, such as click-through rate prediction (click/ignore), medical diagnosis (patient/non-patient), financial fraud detection (fraud/normal transaction), and network intrusion detection (malicious/normal request), etc~\citep{guo2017review-application}.
Imbalanced data often leads to degraded predictive performance of many standard machine learning algorithms since they assume a balanced class distribution and are directly optimized for global accuracy~\citep{he2008review,he2013review}.
Imbalanced learning (IL) aims to tackle the class imbalance problem, i.e., learn an unbiased model from imbalanced data.

Most of the commonly used IL methods are based on resampling and reweighting, which are also the primary interest of existing open-source IL packages such as {\tt imblearn}~\citep{guillaume2017imblearn} and {\tt smote-variants}~\citep{kovacs2019smotevariants}.
Beyond them, ensemble imbalanced learning (EIL) improves typical IL methods by combining the results of multiple independent resampling/reweighting and reducing variance~\citep{galar2012review-ens-imb}.
Recent studies have shown that EIL solutions are highly competitive and gaining increasing popularity in real-world applications \citep{guo2017review-application,dong2020survey}.
However, despite the success of EIL, only a handful of methods are available in existing open-source packages, while many important works \citep{fan1999adacost,chawla2003smoteboost,liu2009easyens-cascade,galar2012review-ens-imb} have no standard Python implementation.

To fill this gap, we present the {\tt imbalanced-ensemble} ({\tt imbens}) Python toolbox to better \textit{leverage the power of ensemble learning to address the class imbalance problem}.
Currently (version 0.2.0), {\tt imbens} have implemented 14 popular EIL methods that based on different imbalance handling strategies (e.g., under/over-sampling and reweighting) and ensemble manners (e.g., boosting and bagging), as summarized in \tablename~\ref{tab:methods}.
These methods are implemented with high-level abstractions based on the taxonomy in \tablename~\ref{tab:methods}, and new algorithms can be easily implemented by taking advantage of inheritance and polymorphism.
{\tt imbens} also provides a number of practical features and utilities for better training, evaluating, and comparing different EIL models.
Altogether, {\tt imbens} is an extensible open-source library for researchers and engineers to handle real-world imbalance learning problems, as well as to develop and benchmark new EIL methods under a standard unified API design.
% The following sections demonstrate the project vision, an overview of included EIL methods, a comparison with existing open-source packages, and the implementation design of {\tt imbens}.
% Finally, we present our conclusion and future plans for the {\tt imbens} package.

\newcommand{\yes}{\ding{51}}
\newcommand{\no}{\ding{55}}

\begin{table}[t]
\begin{threeparttable}
\footnotesize
\resizebox{\linewidth}{!}{
\begin{tabular}{l|ccc|c|c}
\toprule
\multicolumn{1}{c|}{\multirow{2}{*}{Method}}              & \multicolumn{3}{c|}{\textbf{Solution Type}}                                       & \textbf{Ensemble} & \textbf{Multi-} \\ \cline{2-4}
\multicolumn{1}{c|}{}                                     & \multicolumn{1}{c|}{\textbf{US}} & \multicolumn{1}{c|}{\textbf{OS}} & \textbf{RW} & \textbf{Type}     & \textbf{core}   \\ \midrule
{\sc SelfPacedEnsemble}~\citep{liu2020spe}                & \multicolumn{1}{c|}{\yes}        & \multicolumn{1}{c|}{\no}         & \no         & Iterative         & \no             \\
{\sc BalanceCascade}~\citep{liu2009easyens-cascade}       & \multicolumn{1}{c|}{\yes}        & \multicolumn{1}{c|}{\no}         & \no         & Iterative         & \no             \\
{\sc BalancedRandomForest}~\citep{chen2004balance-forest} & \multicolumn{1}{c|}{\yes}        & \multicolumn{1}{c|}{\no}         & \no         & Parallel          & \yes            \\
{\sc EasyEnsemble}~\citep{liu2009easyens-cascade}         & \multicolumn{1}{c|}{\yes}        & \multicolumn{1}{c|}{\no}         & \no         & Parallel          & \yes            \\
{\sc RusBoost}~\citep{seiffert2010rusboost}               & \multicolumn{1}{c|}{\yes}        & \multicolumn{1}{c|}{\no}         & \yes        & Iterative         & \no             \\
{\sc UnderBagging}~\citep{barandela2003underbagging}      & \multicolumn{1}{c|}{\yes}        & \multicolumn{1}{c|}{\no}         & \no         & Parallel          & \yes            \\
{\sc OverBoost}~\citep{galar2012review-ens-imb}           & \multicolumn{1}{c|}{\no}         & \multicolumn{1}{c|}{\yes}        & \yes        & Iterative         & \no             \\
{\sc SmoteBoost}~\citep{chawla2003smoteboost}             & \multicolumn{1}{c|}{\no}         & \multicolumn{1}{c|}{\yes}        & \yes        & Iterative         & \no             \\
{\sc KmeansSmoteBoost}~\citep{chawla2003smoteboost}       & \multicolumn{1}{c|}{\no}         & \multicolumn{1}{c|}{\yes}        & \yes        & Iterative         & \no             \\
{\sc OverBagging}~\citep{wang2009over-smote-bagging}      & \multicolumn{1}{c|}{\no}         & \multicolumn{1}{c|}{\yes}        & \no         & Parallel          & \yes            \\
{\sc SmoteBagging}~\citep{wang2009over-smote-bagging}     & \multicolumn{1}{c|}{\no}         & \multicolumn{1}{c|}{\yes}        & \no         & Parallel          & \yes            \\
{\sc AdaCost}~\citep{fan1999adacost}                      & \multicolumn{1}{c|}{\no}         & \multicolumn{1}{c|}{\no}         & \yes        & Iterative         & \no             \\
{\sc AdaUCost}~\citep{shawe1999adauboost}                 & \multicolumn{1}{c|}{\no}         & \multicolumn{1}{c|}{\no}         & \yes        & Iterative         & \no             \\
{\sc AsymBoost}~\citep{viola2001asymboost}                & \multicolumn{1}{c|}{\no}         & \multicolumn{1}{c|}{\no}         & \yes        & Iterative         & \no             \\ \bottomrule
\end{tabular}
}
\begin{tablenotes}
    \centering
    \tiny
    \item * Abbreviations: under-sampling (US), over-sampling (OS), reweighting (RW), {\tt imblearn} ({\tt imbln}), {\tt smote-variants} ({\tt sv}).
\end{tablenotes}
\end{threeparttable}
\caption{Ensemble imbalanced learning methods implemented in {\tt imbens}.}
\label{tab:methods}
\end{table}
\section{Project Features}

\newcommand{\docurl}{https://imbalanced-ensemble.readthedocs.io}

\textbf{Quality assurance.}
This project follows the {\tt PEP8} standard. 
In order to ensure code quality, a standard set of unit tests is provided leading to a coverage of 96\% for the release 0.2.0 of the toolbox.
To allow both the user and the developer to either use or contribute to this toolbox, \textit{CircleCI} is used to easily integrate new code and ensure back-compatibility.

\noindent
\textbf{Documentation.}
All EIL methods implemented in {\tt imbens} share a unified API design similar to {\tt scikit-learn}~\citep{pedregosa2011sklearn}.
Detailed documentation is developed using {\tt sphinx} and {\tt numpydoc} and rendered using \textit{ReadtheDocs}\footnote{\href{\docurl}{\tt \docurl}}, including comprehensive API references, installation guideline, and code usage examples.

\noindent
\textbf{Functionalities.}
{\tt imbens} provides a number of extended features to ease the usage of EIL algorithms in practice.
With a few parameters, users can easily customize the sampling scheduler/cost-matrix and logging information/granularity, thus gaining more precise training and verbose logging control.
Additionally, a set of utility functions and modules are provided to evaluate, compare, and benchmark different EIL algorithms.

\noindent
\textbf{Openness.}
{\tt imbens} is distributed under the MIT license.
The code repository is hosted on GitHub to facilitate collaboration and documented contribution guidelines are provided.
At the time of this writing, five contributors have participated in the form of bug reports/fixes.

\noindent
\textbf{Project relevance.}
% {\tt imbens} has been used/included in multiple popular open-source projects\footnote{Examples: \href{https://github.com/AutoViML/Auto_ViML}{\tt Auto\_ViML}, \href{https://github.com/AutoViML/featurewiz}{\tt featurewiz}, \href{https://github.com/josephmisiti/awesome-machine-learning}{\tt awesome-machine-learning}, \href{https://github.com/ZhiningLiu1998/awesome-imbalanced-learning}{\tt awesome-imbalanced-learning}}.
{\tt imbens} has been used/included in multiple open-source projects such as \href{https://github.com/AutoViML/Auto_ViML}{\tt Auto\_ViML}, \href{https://github.com/AutoViML/featurewiz}{\tt featurewiz}, and \href{https://github.com/josephmisiti/awesome-machine-learning}{\tt awesome-machine-learning}.
At the edition time, the GitHub repository has received 180+ stars and its PyPI downloads exceed 1,000 per month.

% \noindent
% \textbf{Extended functionalities.}
% Most existing EIL methods are designed for binary imbalanced classification.
% We extend their design in {\tt imbens} to support multi-class imbalanced learning, allowing them to be employed in a wider range of applications.
% We also provide additional options such as customizable resampling scheduler for more fine-grained training control.

% \noindent
% \textbf{Customizable logging and visualization.}
% {\tt imbens} provides easy access to the status and statistics of the ensemble training process.
% With a few parameters, users can easily customize the information they want to monitor during training, including evaluation datasets, metrics, and log granularity.
% We also implement a general ensemble visualizer to provide further information and/or make comparison between multiple classifiers.

% \noindent
% \textbf{Performance and compatibility.}
% Parallelization is enabled for both resampling and ensemble training when possible.
% The implemented EIL classifiers, visualizer, and utilities are also fully compatible with other popular packages like {\tt scikit-learn} and {\tt imblearn}.
\section{Library Design and Implementation}

% \noindent
% \textbf{Included methods.}
% Currently (version 0.2.0), {\tt imbens} have implemented 14 popular EIL methods, as summarized in \tablename~\ref{tab:methods}.
% Their IL solutions can be divided into two main groups: resampling (under/over-sampling) and reweighting  (boosting/cost-sensitive learning).
% Note that some of them combine resampling and reweighting, e.g., {\sc SmoteBoost}.
% These methods also involve two different ensemble training manners: iterative (e.g., boosting) and parallel (e.g., bagging).
% Multi-core parallelization is enabled for all parallel EIL methods in {\tt imbens}.
% Up to our knowledge, we provide the first standard Python implementation for 10 of the 14 included EIL methods.
% For the techniques that are included in existing packages (e.g., {\tt imblearn}), we also provide extended functionalities with additional parameters (\tablename~\ref{tab:fit-params}) for more precise training control and customizable verbose logging.

% Up to our knowledge, we provide the first standard Python implementation for 10 of the 14 included EIL methods.
% Although the other 4 can be implemented with the {\tt imblearn} package, they lack many of the useful features from {\tt imbens} such as sampling scheduler and dynamic training logs.
% The {\tt smote-variants} package focuses only on resampling techniques, especially oversampling, and does not involve any ensemble learning approaches.

\begin{table}[t]
\centering
\scriptsize
\begin{threeparttable}
\resizebox{\linewidth}{!}{
\begin{tabular}{c|c|c|c}
\toprule
\textbf{Parameter} & \textbf{Data Type} & \textbf{Availability} & \textbf{Description} \\
\midrule
{\tt target\_label}       & int            & RS     & Specify the class targeted by the resampling.           \\
{\tt n\_target\_samples}  & int/dict       & RS     & Specify the desired number of samples (of each class).  \\
{\tt balancing\_schedule} & str/callable   & RS+IT  & Scheduler that controls resampling during the training. \\
{\tt cost\_matrix}        & str/array      & CS     & Specify (how to set) the misclassification cost matrix. \\
{\tt eval\_datasets}      & dict           & All    & Dataset(s) used for evaluation during the training.     \\
{\tt eval\_metrics}       & dict           & All    & Metric(s) used for evaluation during the training.      \\
{\tt train\_verbose}      & bool/int/dict  & All    & Controls the verbosity during ensemble training.        \\ 
\bottomrule
\end{tabular}
}
\begin{tablenotes}
    \centering
    \tiny
    \item * Abbreviations: resampling (RS), cost-sensitive (CS), iterative ensemble (IT).
\end{tablenotes}
\end{threeparttable}
\caption{Additional key parameters of the {\tt fit} method in {\tt imbens}.}
\label{tab:fit-params}
\end{table}

% \noindent
% \textbf{API design.}
The {\tt imbens} package relies on {\tt numpy}, {\tt pandas}, {\tt scipy}, and {\tt scikit-learn}.
We use {\tt joblib} to implement multi-core execution for supported algorithms.
Inspired by {\tt scikit-learn}'s API design \citep{sklearn_api}, all EIL algorithms inherit from a base class and share the same interface: (i) {\tt fit} builds an ensemble classifier from the class-imbalanced training set $(X,Y)$; (ii) {\tt predict} returns the predicted class labels corresponding to the given input samples; and (iii) {\tt predict\_proba} gives predicted class probabilities instead of labels.
Additionally, inspired by {\tt imblearn}'s API design \citep{guillaume2017imblearn} and the taxonomy in \tablename~\ref{tab:methods}, we decouple the implementation of imbalance handling and ensemble training in {\tt imbens}.
For example, all resampling + boosting EIL models (e.g., {\sc RusBoost}, {\sc SmoteBoost}) inherit the {\tt ResampleBoostClassifier}, only with different samplers (e.g., {\tt RandomUnderSampler}, {\tt SMOTE}) from the {\tt imbens.samplers} module.
New models can be easily implemented within this framework by taking advantage of inheritance and polymorphism.

All EIL models take two key parameters for initialization: {\tt estimator} and {\tt n\_estimators}.
The former can be any {\tt scikit-learn}-style classifier instance, and the latter is an integer that specifies the size of the ensemble.
To enable more precise control and monitoring of the EIL training process, the {\tt fit} function takes several additional keyword arguments.
{\tt target\_label}, {\tt n\_target\_samples} and {\tt balancing\_schedule} can be used to dynamically adjust the sampling strategy during training, and {\tt cost\_matrix} allows the user to specify the misclassification cost for each class.
Besides, {\tt eval\_datasets}, {\tt eval\_metrics}, and {\tt train\_verbose} control the content and granularity of the ensemble training log.
\tablename~\ref{tab:fit-params} summarizes the data type, availability, and semantics of these keyword arguments.

\begin{figure}[h]
    \centering
    \subfigure[{\tt confusion\_matrix\_heatmap()}]{
    \includegraphics[width=0.47\linewidth]{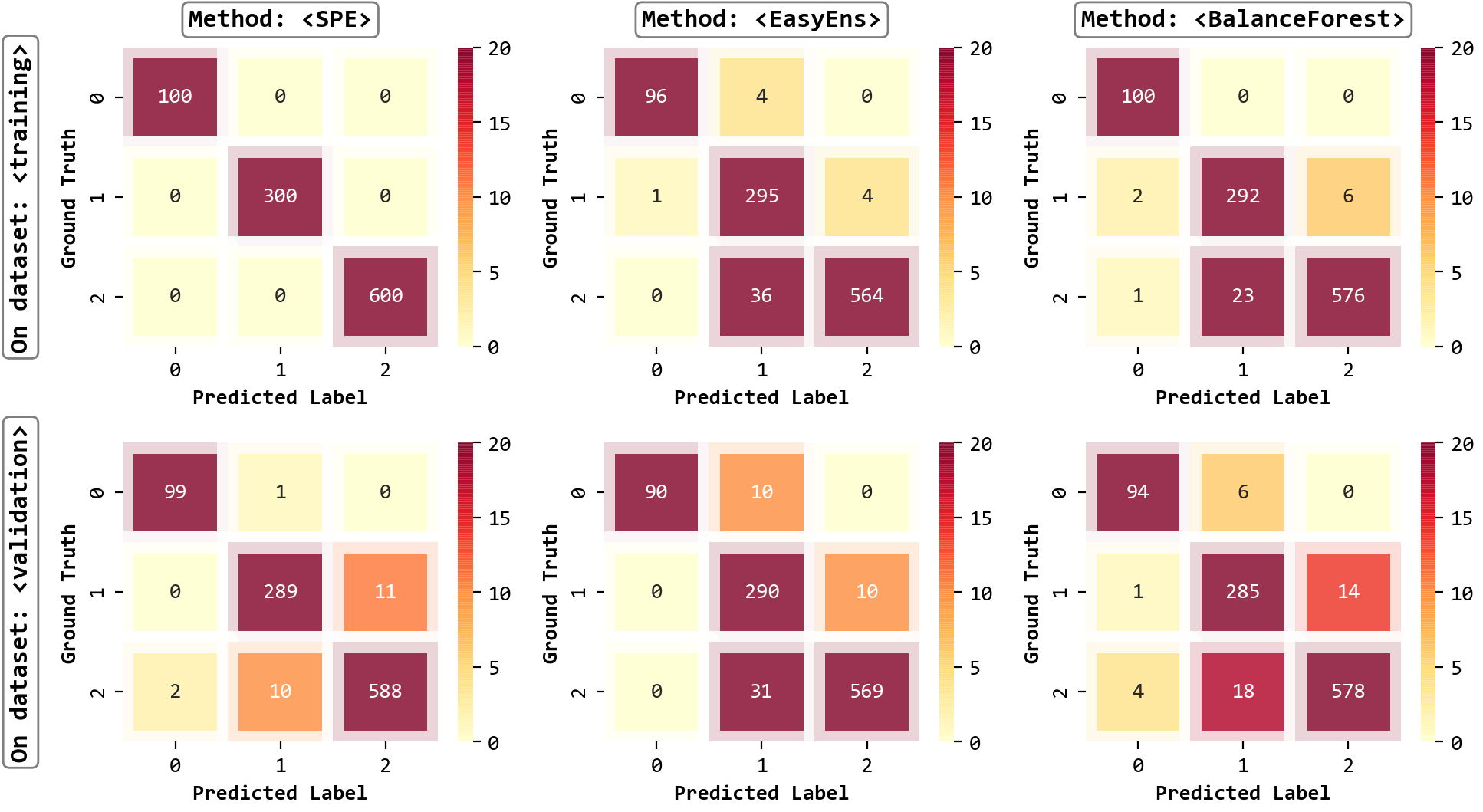}
    }
    \subfigure[{\tt performance\_lineplot()}]{
    \includegraphics[width=0.47\linewidth]{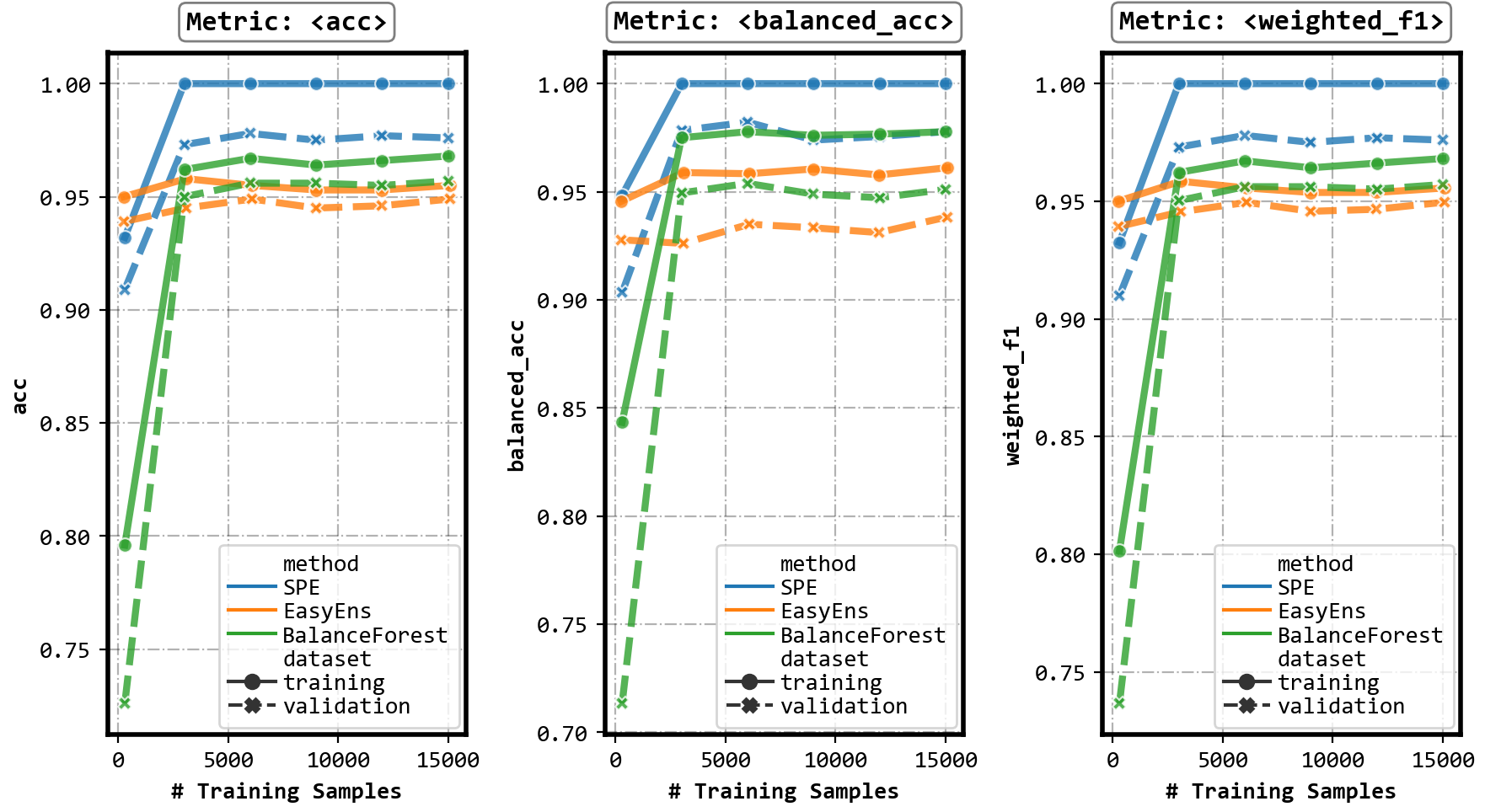}
    }
    \caption{Examples of using {\tt ImbalancedEnsembleVisualizer} for visualization.}
    \label{fig:visualize}
\end{figure}

Additionally, {\tt imbens} provides a set of utility functions ({\tt generate\_imbalance\_data}, {\tt evaluate\_print}, etc.) in the {\tt imbens.utils} and {\tt imbens.visualizer} modules.
With these utilities, Users can easily create synthetic imbalanced datasets, test EIL models, and (visually) evaluate and compare multiple EIL models.
Code Snippet 1 is a demo showcasing how the deployment, evaluation, and visualization of EIL models can be conveniently conducted using the {\tt imbens} API.
\figurename~\ref{fig:visualize} shows examples of evaluating various EIL models on multiple datasets and metrics using the {\tt ImbalancedEnsembleVisualizer}\footnote{More examples can be found in the \href{https://imbalanced-ensemble.readthedocs.io/en/latest/auto_examples/index.html}{{\tt imbens} documentation}.}.
% Visualization examples provided by the {\tt ImbalancedEnsembleVisualizer} are shown in \figurename~\ref{fig:visualize}.
% More examples can be found in the imbens gallery.

\vspace{2pt}
\begin{lstlisting}[title={Code Snippet 1: Demo of {\tt imbens} API with the {\sc SelfPacedEnsemble} classifier.},captionpos=b]
  >>> from imbens import ensemble, datasets, utils, visualizer
  >>> 
  >>> X_train, X_test, y_train, y_test = datasets.generate_imbalance_data(\
  ...     n_samples=200, weights=[.9,.1], test_size=.5)      # prepare data
  >>> 
  >>> clf = ensemble.SelfPacedEnsembleClassifier()           # initialize ensemble
  >>> clf.fit(X_train, y_train)                              
  >>> y_test_pred = clf.predict(X_test)                      # predict labels
  >>> utils.evaluate_print(y_test, y_test_pred, "SPE")       # performance evaluation
  SPE balanced Acc: 0.972 | macro Fscore: 0.886 | macro Gmean: 0.972
  >>> 
  >>> vis = visualizer.ImbalancedEnsembleVisualizer()        # initialize visualizer
  >>> vis.fit({'SPE': clf})
  >>> vis.performance_lineplot()                             # performance visualization 
  >>> vis.confusion_matrix_heatmap()                         # prediction visualization
\end{lstlisting}

\section{Conclusion and future plans}

In this paper, we present {\tt imbens}, a comprehensive Python toolbox for out-of-the-box ensemble class-imbalanced learning.
As avenues for future work, we plan to include additional ensemble imbalanced learning methods that are based on evolutionary algorithm/meta-learning/hybrid-sampling, as well as more detailed examples, user guides and tutorials.

% Acknowledgements should go at the end, before appendices and references

\acks{The authors would like to thank (i) Zhepei Wei, Erxin Yu, Qiang Huang, Kai Guo, Boyang Yu, Zhaonian Cai, Hangting Ye from Jilin University, (ii) Wei Cao, Jiang Bian from Microsoft Research, (iii) Pengfei Wei from ByteDance Singapore, and (iv) Jing Jiang from University of Technology Sydney, for their valuable suggestions and comments during the development of this project.}

% Manual newpage inserted to improve layout of sample file - not
% needed in general before appendices/bibliography.

% \input{sections/appendix}

\vskip 0.2in
\bibliography{reference}

\end{document}